\pdfoutput=1
\documentclass[11pt]{article}

\usepackage[preprint]{acl}

\usepackage{times}
\usepackage{latexsym}
\usepackage{paralist}

\usepackage[T1]{fontenc}
\usepackage[utf8]{inputenc}
\usepackage{microtype}
\usepackage{inconsolata}

\usepackage{graphicx}
\usepackage{booktabs}
\usepackage{multirow}
\usepackage{adjustbox}
\usepackage{soul}
\usepackage{amsmath}
\usepackage[capitalise]{cleveref}
\usepackage{arydshln}

\usepackage{bbm}

\usepackage[most]{tcolorbox}
\usepackage{enumitem}
\usepackage{titlesec}
\usepackage{lipsum}

\usepackage{tcolorbox}
\usepackage{hyperref}
\usepackage{etoolbox}

\usepackage{caption}
\definecolor{darkgreen}{RGB}{0,100,0}
\usepackage{authblk}

\newcommand{\ourname}[1]{CLATTER}{}

\title{CLATTER: Comprehensive Entailment Reasoning for Hallucination Detection}

\author{
    \textbf{Ron Eliav$^{1}$} \qquad
    \textbf{Arie Cattan$^{1}$} \qquad
    \textbf{Eran Hirsch$^{1}$} \qquad
    \textbf{Shahaf Bassan$^{2}$} \\[0.7em]
    \textbf{Elias Stengel-Eskin$^{3}$} \qquad
    \textbf{Mohit Bansal$^{3}$} \qquad
    \textbf{Ido Dagan$^{1}$}
}
\affil{$^{1}$Bar-Ilan University \quad{} $^{2}$Hebrew University of Jerusalem \quad{} $^{3}$UNC Chapel Hill}
\affil{\tt roneliav1@gmail.com} 

\begin{document}
\maketitle

\begin{abstract}
A common approach to hallucination detection casts it as a natural language inference (NLI) task, often using LLMs to classify whether the generated text is entailed by corresponding reference texts. 
Since entailment classification is a complex reasoning task, one would expect that LLMs could benefit from generating an explicit reasoning process, as in CoT reasoning or the explicit ``thinking'' of recent reasoning models.
In this work, we propose that guiding such models to perform a systematic and comprehensive reasoning process---one that both decomposes the text into smaller facts and also finds evidence in the source for each fact---allows models to execute much finer-grained and accurate entailment decisions, leading to increased performance. 
To that end, we define a 3-step reasoning process, consisting of (i) claim decomposition, (ii) sub-claim attribution and entailment classification, and (iii) aggregated classification, showing that such guided reasoning indeed yields improved hallucination detection.
Following this reasoning framework, we introduce an analysis scheme, consisting of several metrics that measure the quality of the intermediate reasoning steps, which provided additional empirical evidence for the improved quality of our guided reasoning scheme.

\end{abstract}

\section{Introduction}
\label{sec:intro}

\begin{figure}[!t]
    \centering
    \includegraphics[width=\columnwidth]{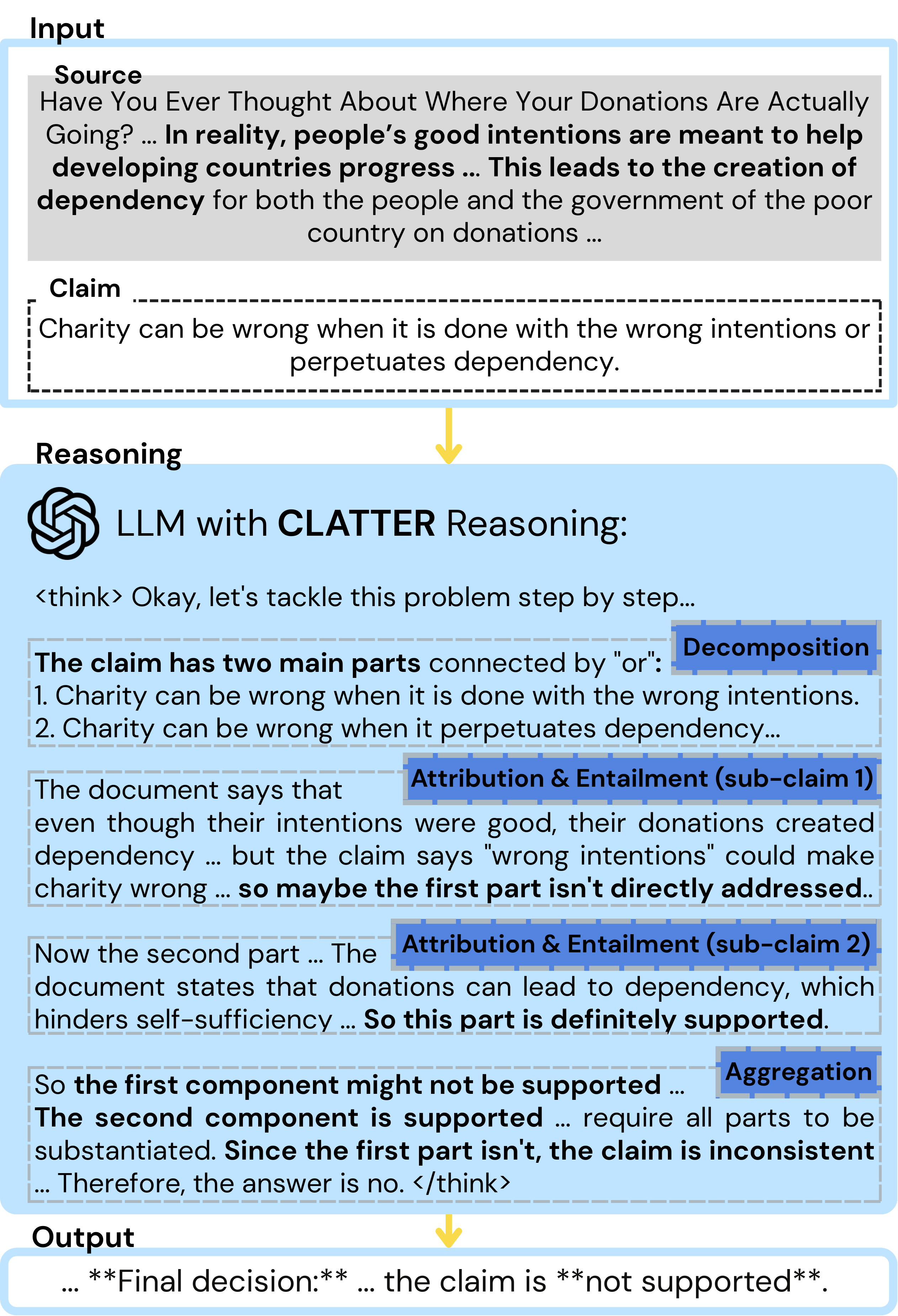}
    \caption{An example of \ourname{} reasoning framework to evaluate a claim. The process begins by decomposing the claim into its two sub-claims. Each sub-claim is checked against the source via attribution and entailment analysis. Finally, the results are aggregated to reach a not supported verdict for the overall claim.}
    \label{fig:real_reasoning_example}
\end{figure}

The output of Large Language Models (LLMs) is often required to be faithful to some reference texts. Such texts might be provided by the user, as in text summarization, retrieved sources, as in RAG settings, or retrieved references against which parametric-based generation is verified for factuality. 
In such settings, a critical challenge is to detect if the generated output contains unsupported claims, known as hallucinations \citep{tian2020stickingfactsconfidentdecoding,thorat2024summexeceditfactualconsistencybenchmark,kovács2025lettucedetecthallucinationdetectionframework,paudel2025hallucinothallucinationdetectioncontext}. 
Automated hallucination detection methods can inform users of suspected hallucinations \citep{10.1145/3613904.3642428,zhao-etal-2024-successfully}, correct hallucinations by editing the output \citep{wadhwa-etal-2024-learning-refine}, or guide models to avoid hallucinations through reinforcement learning \citep{roit-etal-2023-factually} and controlled decoding \citep{wan-etal-2023-faithfulness}. 

\begin{figure*}[t]
    \centering
    \includegraphics[width=\textwidth]{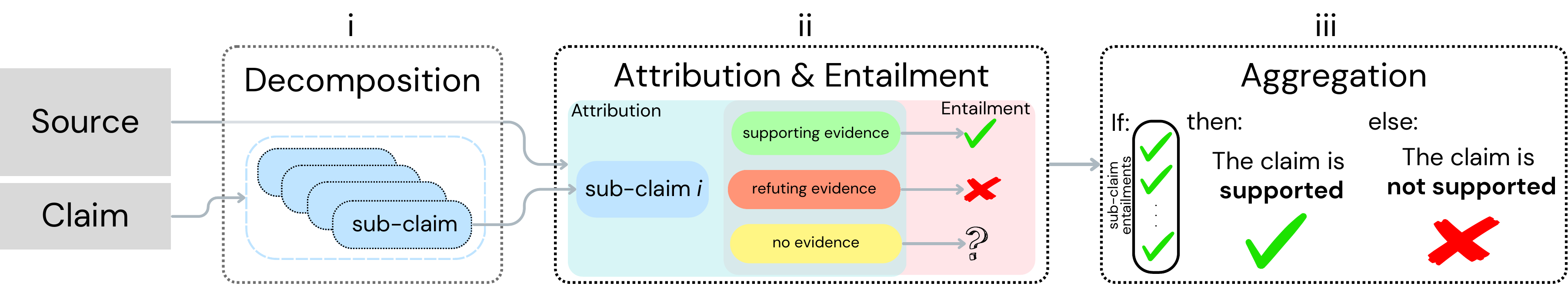}
    \caption{Overview of \ourname{} process. (i) Decomposition: the original claim is split into individual sub-claims. (ii) Attribution \& Entailment: each sub-claim is checked against the source for supporting evidence, refuting evidence, or no evidence. (iii) Aggregation: if all sub-claims are supported, the claim is accepted; otherwise, it is rejected.}
    \label{fig:xnli_approach}
\end{figure*}

The task of hallucination detection is mostly seen as an entailment classification task \cite{10.1007/11736790_9,bowman-etal-2015-large}, where the hypothesis is a model-generated output claim while the premise is the source text. Hallucination detection is then implemented using either fine-tuned entailment classifiers \citep{zha-etal-2023-alignscore, kamoi2023wice, tang-etal-2024-minicheck}, or via prompting LLMs to complete the entailment task \citep{kamoi2023wice, laban-etal-2023-summedits, min-etal-2023-factscore, tang-etal-2024-tofueval}. 
In our work we focus on the latter scenario, where LLMs are often preferred thanks to their broad domain and language coverage, robustness and accessibility. 

Since entailment classification is a complex reasoning task, we expect that LLMs might benefit from generating an explicit reasoning process, as in CoT reasoning or the explicit ``thinking'' of recent reasoning models (Large Reasoning Models, or LRMs). Given such model-generated entailment reasoning, two research questions arise:
\textbf{\emph{RQ1:}} How well do models perform such reasoning on their own, in an un-guided manner? This question is posed with respect to both bottom line entailment classification performance as well as the validity of the reasoning process itself.
\textbf{\emph{RQ2:}} Is it possible to improve such reasoning, by guiding models to perform systematic reasoning steps that follow the inherent semantics of entailment decision-making?

Toward addressing these questions, we first formulate a systematic and comprehensive reasoning process for entailment classification,
which we term \ourname{}: \textbf{C}laim \textbf{L}ocalization \& \textbf{ATT}ribution for \textbf{E}ntailment \textbf{R}easoning. 
This process consists of three steps, namely (i) claim decomposition, (ii) sub-claim attribution and entailment classification, and (iii) aggregated classification, as illustrated in Figures~\ref{fig:real_reasoning_example} and~\ref{fig:xnli_approach}.
Further, we define a set of metrics that measure the validity of the different steps involved in such entailment reasoning.
While prior work also decomposes entailment reasoning based on sub-claims, to the best of our knowledge, we are the first to investigate a principled decomposition of this sort as a single LLM reasoning process, as opposed to prior pipeline architectures, which often involve targeted fine-tuned models \citep{kamoi2023wice, manakul-etal-2023-selfcheckgpt, min-etal-2023-factscore}.

Our experiments show that \ourname{}-guided reasoning does improve bottom-line entailment classification, relative to un-guided reasoning. 
Importantly, \ourname{}-guided LRMs perform sub-claim attribution and entailment classification much more accurately, successfully following the prescribed reasoning steps.

Overall, our contributions include: (1) introducing \ourname{} as a comprehensive multi-step reasoning process for entailment classification by LLMs (Section~\ref{sec:definition}); (2) defining assessment metrics 
for the involved reasoning steps (Section \ref{sec:metrics}); 
(3) analyzing both unguided and \ourname{}-guided reasoning, in both CoT and LRM settings, showing the advantages of \ourname{} reasoning in both entailment classification and reasoning quality.

In the following sections, we describe the \ourname{} approach in detail (\S\ref{sec:definition}), present evaluation metrics for the entailment reasoning steps (\S\ref{sec:metrics}), describe our experimental setup (\S\ref{sec:nli_experiment}), present our results and ablations (\S\ref{sec:results}), discuss insights from our manual analysis (\S\ref{sec:analysis}), and finally contrast with related work (\S\ref{sec:related_work}).

\section{Comprehensive NLI Reasoning}
\label{sec:definition}

In the following section, we formulate the \ourname{} reasoning process, which, in our setting, models are instructed to follow when making an entailment decision.
We take the view that a natural-language sentence can be presented as a conjunction of smaller facts \cite{davidson1967logical, partee2008compositionality}, all sharing a consistent interpretation, where the sentence is semantically equivalent to the union of these facts. Then, a hypothesis is \textit{entailed} if all its facts are entailed by the source, \textit{contradicted} if at least one is contradicted, and \textit{neutral} otherwise.
Consequently, for detecting a hallucination in a given claim, we first decompose a claim into sub-claims. Each sub-claim is then classified by checking for a corresponding piece of evidence in the source: entailed if supported, contradicted if opposed, and neutral if no match is found. Finally, we aggregate the decisions of each sub-claim to provide a prediction for the whole claim.

We propose guiding models to follow a systematic process aligned with this perspective. As shown in \cref{fig:xnli_approach}, the entailment prediction of a generated claim $\mathcal{H}$ relative to a source $\mathcal{P}$ involves three steps: \begin{inparaenum}[(i)] \item decomposition, \item attribution and entailment classification, and \item aggregation. \end{inparaenum}
Through the reasoning process, \ourname{} provides a set of triples ${(h_i, p_i, \hat{y}_i)}$, where $h_i$ is a sub-claim, $p_i$ is the corresponding attribution in the source, and $\hat{y}_i$ denotes the entailment status of $h_i$ relative to $\mathcal{P}$. Finally, \ourname{} aggregates all $\hat{y}_i$ values and returns a final prediction $\hat{y}$ of either \emph{supported} or \emph{not supported}. A detailed explanation of each step is provided below.

\paragraph{(i) Decomposition:} 

The first step in \ourname{} process includes the decomposition of $\mathcal{H}$ into sub-claims. A sub-claim $h_i$ is both entailed by $\mathcal{H}$ and has a verifiable truth value against the source $\mathcal{P}$. For a complete decomposition, the union of all the sub-claims should be semantically equivalent to the full hypothesis. Formally, $\bigcup_i h_i=\mathcal{H}$.
In \cref{fig:real_reasoning_example}, the model decomposes the claim into two parts: \emph{``Charity can be wrong when it is done with the wrong intentions''} and \emph{``Charity can be wrong when it perpetuates dependency.''}

\paragraph{(ii) Attribution \& Entailment:}
In the second step, the model looks for evidence and determines the entailment for each sub-claim $h_i$.
\begin{inparaenum}[(a)] 
\item \textbf{Attribution}: Search the source text for an evidence $p_i \in \mathcal{P}$ that is either entailing (supporting) or contradicting (refuting) the sub-claim.
\item \textbf{Entailment}: If supporting or refuting evidence is found, classify the sub-claim accordingly. Otherwise, classify it as neutral.
\end{inparaenum} In step `Attribution \& Entailment (sub-claim 2)' in Figure \ref{fig:real_reasoning_example}, a supporting attribution is found, leading to an \emph{entailment} classification of this sub-claim.

\paragraph{(iii) Aggregation:} 

In the final step, the model aggregates the entailment labels of the sub-claims following the logic: if all sub-claims are \textit{entailed}, the claim is \textit{supported}; otherwise, the claim is \textit{not-supported}. For example, in \cref{fig:real_reasoning_example}, one sub-claim is \emph{neutral}, therefore the claim is \emph{not-supported}.

Overall, these three steps combine the decomposition of the full semantics of a claim into sub-claims, the verification of the entailment of each sub-claim, and the aggregation of all decisions. All in one reasoning process. This flow makes \ourname{} approach both \emph{comprehensive} and \emph{systematic}. The full instructions provided to the models are listed in Appendix~\ref{app:prompts}.

\section{Evaluation Metrics for Entailment Reasoning}
\label{sec:metrics}

As discussed in \cref{sec:intro}, two of our objectives are to analyze the innate reasoning produced by LRMs and the ability of LRMs to follow \ourname{} instructions.
Inspired by the components of the \ourname{} process, we propose to assess entailment reasoning steps by three corresponding components (decomposition, attribution \& entailment, and aggregation). Additionally, in Section \ref{sec:analysis} we show that these metrics are instruction-agnostic and are relevant for instruction-free reasoning as well as other reasoning for NLI.
To compute the metrics, we assume the ability to extract sub-claims, attribution, entailment labels, and the final decision from the model's reasoning. As LRMs express reasoning in natural language, this extraction is non-trivial. Instead of relying on potentially noisy automated metrics, we opt to analyze and score these metrics manually, thus ensuring the quality of our results.

\paragraph{Atomicity.}

Following \ourname{}, models are instructed to decompose a hypothesis into sub-claims during reasoning. We define the \emph{atomicity} metric to capture this behavior. 
\citet{wanner-etal-2024-closer} proposed to count the number of sub-claims produced by a decomposer as part of the decomposer evaluation. Similarly, we suggest counting the number of distinct sub-claims $\mathcal{H} = \{h_1, h_2, \dots, h_n\}$ generated at the decomposition step. If no decomposition occurs, $\mathcal{H}$ contains a single element. The atomicity score is then defined as: $A_{\text{tomicity}} := |\mathcal{H}|$. This metric has no ground-truth value, but it can influence later steps. Low atomicity leads to longer and more complex sub-claims, making attribution and entailment classification harder. High atomicity increases the risk of unfaithful or incomplete decompositions.

\paragraph{Soundness.}

As part of the decomposition step, we assess whether the model, in its reasoning steps, generates sub-claims that are not semantically entailed by the claim. The \emph{soundness} metric measures the proportion of generated sub-claims that are consistent with the claim. The soundness score is defined as:

\begin{equation}
S_{\text{oundness}} := \frac{1}{|\mathcal{H}|} \sum_{i=1}^{|\mathcal{H}|} \mathbbm{1}_{\{h_i \text{ is sound}\}}
\end{equation}

\noindent Intuitively, a low soundness score suggests the model introduces extraneous or fabricated sub-claims during decomposition, risking incorrect entailment judgments.

\paragraph{Completeness.}

For a complete view of the decomposition step, we evaluate whether the model refers all the semantic content of the original claim. The \emph{completeness} metric checks if any part was omitted during decomposition. It is a binary value: $1$ if all information is covered by the model’s sub-claims, and $0$ if any is missing. The completeness score is then defined as:

\begin{equation}
C_{\text{ompleteness}} :=
\begin{cases}
1 & \text{if } \mathcal{H} \subseteq \bigcup_i h_i  \\
0 & \text{otherwise}
\end{cases}
\end{equation}

\noindent Intuitively, this metric highlights cases where the model omits parts of the claim---especially contradicting ones---potentially leading to incorrect predictions like falsely labeling it as \textit{entailed}.

\paragraph{Sub-claim Attribution.}
The first phase in the second component of \ourname{} is the attribution for each sub-claim. The \emph{attribution} metric assesses whether the model correctly identifies supporting or contradicting evidence from the source for each sub-claim, when such evidence exists. An attribution is correct if it can justify the entailment label of the sub-claim. Additionally, if the model does not find any evidence in the source when no such evidence exists, the model receives a full score on this sub-claim.
\begin{equation}
A_{\text{ttribution}} := \frac{1}{|\mathcal{H}|} \sum_{i=1}^{|\mathcal{H}|} \mathbbm{1}_{\{h_i \text{ is correctly attributed}\}}
\end{equation}

\noindent Intuitively, incorrect or missing attribution can cause sub-claim misclassification, leading to an incorrect overall entailment decision.

\paragraph{Sub-claim Entailment Classification.}

The second phase in `Attribution \& Entailment' step is to determine the entailment classification of each sub-claim. The \emph{entailment} metric evaluates whether the model correctly predicts the entailment label for each sub-claim, comparing the predicted label $\hat{y}_i$ with the gold $y_i$ given by an oracle (or by a human evaluator).\footnote{For a binary classification, the \emph{neutral} and \emph{contradicted} classes may be grouped under a single \emph{not supported} class.} The entailment metric is defined by:
\begin{equation}
E_{\text{ntailment}} := \frac{1}{|\mathcal{H}|} \sum_{i=1}^{|\mathcal{H}|} \mathbbm{1}_{\{\hat{y}_i = y_i\}}
\end{equation}

\noindent Intuitively, misclassifying even \emph{one} sub-claim can impact the overall claim prediction, making this step crucial for performance.

\paragraph{Aggregation.}
Finally, for the last step of \ourname{}, we assess whether the model correctly aggregates sub-claim entailment predictions into a final global decision for the full claim. The \emph{aggregation} metric follows this logic: \begin{inparaenum}[(i)] \item If all sub-claims are entailed, the hypothesis is \emph{supported}; \item Otherwise, it is classified as \emph{not supported}.
\end{inparaenum}

Let $\hat{y}_{\text{global}}$ be the model’s final prediction for the whole claim, and let $f(\hat{y}_1, \dots, \hat{y}_{|\mathcal{H}|})$ denote the correct aggregated label based on the sub-claim predictions. The aggregation metric is defined as:

\begin{equation}
A_{\text{ggregation}} := \mathbbm{1}_{\{\hat{y}_{\text{global}} = f(\hat{y}_1, \dots, \hat{y}_{|\mathcal{H}|})\}}
\end{equation}

\noindent Intuitively, this binary metric is $1$ if the model's global decision matches the logical aggregation of sub-claim labels, and $0$ otherwise. It captures cases where sub-claim entailment decisions are correct, but the final decision misapplies the aggregation logic.

\section{Experimental Setup}
\label{sec:nli_experiment}

In this section, we describe the experimental setup for hallucination detection, including the methods, datasets, and models used.
The complete prompt templates for all the following approaches are included in \cref{app:prompts}. Experimental results and analysis are presented in \cref{sec:results}.

\subsection{Methods for NLI}

This setup mainly includes the approaches to reasoning about entailment decisions. Our experiment compares several approaches to reasoning for the entailment task. Therefore, all of these approaches are implemented as different reasoning processes for LLMs.

\noindent (1) As a \textbf{baseline} approach, we instruct the model to assess whether a given hypothesis is factually consistent with a provided source, without any instruction on how to make this decision.

\noindent (2) \textbf{\ourname{}}: In our proposed approach, we direct the model to perform systematic and comprehensive reasoning before the entailment decision, as detailed in Section~\ref{sec:definition} and presented in \cref{fig:xnli_approach}.

In addition, for a complete comparison, we add a comparison of one more approach for the entailment task:

\noindent (3) \textbf{QA-Based}: Inspired by prior work using QA pairs for semantic representation and faithfulness verification \citep{he2015question,klein2022qasem,Cattan2024LocalizingFI,dhuliawala-etal-2024-chain}, we instruct the model to generate questions from the hypothesis, answer them using both the hypothesis and the source, and assess entailment via answer equivalence. See Appendix \ref{app:nli_comparison} for details.

\subsection{Datasets}
Numerous datasets have recently been developed for the NLI task. In our study, we focus on three prominent domains:
(1) \textit{Fact Verification}, where a factual claim is verified against a source;
(2) \textit{Question Answering}, where an answer is verified against a set of retrieved passages; and
(3) \textit{Summarization}, where the faithfulness of a summary is evaluated relative to the source document.

To ensure specialization in hallucination detection, we selected one dataset from each domain in which the statements to be evaluated are generated by LLMs.
For the fact verification domain, we use the ClaimVerify dataset \cite{liu-etal-2023-evaluating}. In the question answering domain, we evaluate on the LFQA-Verification dataset \cite{chen2023understanding}. For summarization, we use the TofuEval dataset \cite{tang-etal-2024-tofueval} based on the MediaSum benchmark \cite{zhu-etal-2021-mediasum}. Further details on the subset we chose are presented in \cref{app:nli_experiment}. In our framework, a model is given a source and a generated claim, and should provide a prediction whether the given claim is faithful, relative to the source, or not (i.e., contains hallucination).

\subsection{Models}

We conduct an extensive investigation on four LRMs, instructing them to follow \ourname{} principles. The models evaluated include: \texttt{QwQ-32B-Preview} \cite{qwq-32b-preview}, \texttt{DeepSeek-R1}\cite{guo2025deepseek}, \texttt{O4-mini} \citep{openai2025o3o4mini}, and \texttt{Gemini-2.5-Pro} \citep{google_gemini_2.5_pro}.

As a baseline, we also apply the same process to non-reasoning models---standard LLMs that were not explicitly trained to generate intermediate reasoning before making predictions. This allows us to compare the effectiveness of \ourname{} across both model types and assess whether reasoning-trained models benefit more from structured instruction than standard LLMs.
For non-reasoning models, we evaluate \texttt{Qwen-Plus} \citep{alibaba_qwen_plus}, \texttt{DeepSeek-V3} \cite{deepseek_v3}, \texttt{GPT-4o-mini} \citep{openai_gpt4o}, and \texttt{Gemini-2.0-Flash} \cite{google_gemini_2_0_flash}.
We also report results for the MiniCheck model to provide a comparison with a state-of-the-art fine-tuned baseline.

\section{Results}
\label{sec:results}

We divide our results into two sections. The first is a comparison between the baseline approach and \ourname{} approach. The second is a comparison between the two instruction approaches suggested above (\cref{sec:nli_experiment}: QA-based, and \ourname{}). The results for the former are presented in \cref{subsec:main_results}, and the latter results are presented in \cref{app:nli_experiment}. 
In addition, we conduct an ablation study of each component in the proposed comprehensive instruction, which is detailed in \cref{subsec:ablations}.

\subsection{Entailment Classification Results}
\label{subsec:main_results}
\begin{table*}[t]
\centering
\small
\adjustbox{max width=\textwidth}{%
\begin{tabular}{llccc|ccc|ccc|c}
\toprule
& \textbf{Model} 
& \multicolumn{3}{c|}{\textbf{ClaimVerify}} 
& \multicolumn{3}{c|}{\textbf{LFQA}} 
& \multicolumn{3}{c|}{\textbf{TofuEval}}
& \textbf{Avg} \\
& & Baseline & \ourname{} & $\Delta$ 
& Baseline & \ourname{} & $\Delta$ 
& Baseline & \ourname{} & $\Delta$ 
& $\Delta$ \\
\midrule
\parbox[t]{3mm}{\multirow{1}{*}{\rotatebox[origin=c]{90}{FT}}} &  MiniCheck     & 60.20 & --    & --                         & 55.60 & --    & --                         & 66.20 & --    & --                         & -- \\
\midrule
\parbox[t]{3mm}{\multirow{4}{*}{\rotatebox[origin=c]{90}{LLM}}} & Qwen-Plus     & 71.00 & \textbf{74.40} & \textcolor{darkgreen}{↑~3.40}  & 79.60 & \textbf{81.00} & \textcolor{darkgreen}{↑~1.40}  & \textbf{78.60} & 71.40 & \textcolor{red}{↓~7.20}  & \textcolor{red}{↓~0.80} \\
& Deepseek-V3   & 66.60 & \textbf{73.40} & \textcolor{darkgreen}{↑~6.80}  & 80.60 & \textbf{84.00} & \textcolor{darkgreen}{↑~3.40}  & \textbf{77.80} & 77.20 & \textcolor{red}{↓~0.60}  & \textcolor{darkgreen}{↑~3.20} \\
& GPT-4o-mini   & 71.40 & \textbf{73.80} & \textcolor{darkgreen}{↑~2.40}  & 77.60 & \textbf{83.20} & \textcolor{darkgreen}{↑~5.60}  & \textbf{79.00} & 78.00 & \textcolor{red}{↓~1.00}  & \textcolor{darkgreen}{↑~2.33} \\
& Gemini-2.0    & 68.00 & \textbf{75.00} & \textcolor{darkgreen}{↑~7.00}  & 78.20 & \textbf{80.60} & \textcolor{darkgreen}{↑~2.40}  & \textbf{78.60} & 78.20 & \textcolor{red}{↓~0.40}  & \textcolor{darkgreen}{↑~3.00} \\
\midrule
\parbox[t]{3mm}{\multirow{4}{*}{\rotatebox[origin=c]{90}{LRM}}} &  QwQ-32B-Preview       & 67.40 & \textbf{72.40} & \textcolor{darkgreen}{↑~5.00}  & 79.80 & \textbf{82.40} & \textcolor{darkgreen}{↑~2.60}  & 70.22 & \textbf{79.80} & \textcolor{darkgreen}{↑~9.58}  & \textcolor{darkgreen}{↑~5.72} \\
& DeepSeek-R1   & 69.60 & \textbf{75.60} & \textcolor{darkgreen}{↑~6.00}  & 80.60 & \textbf{84.40} & \textcolor{darkgreen}{↑~3.80}  & 71.23 & \textbf{77.00} & \textcolor{darkgreen}{↑~5.77}  & \textcolor{darkgreen}{↑~5.19} \\
& O4-mini       & 73.20 & \textbf{80.20} & \textcolor{darkgreen}{↑~7.00}  & 85.80 & \textbf{86.80} & \textcolor{darkgreen}{↑~1.00}  & 80.20 & \textbf{81.60} & \textcolor{darkgreen}{↑~1.40}  & \textcolor{darkgreen}{↑~3.13} \\
& Gemini-2.5    & 73.40 & \textbf{76.20} & \textcolor{darkgreen}{↑~2.80}  & \textbf{85.80} & 84.00 & \textcolor{red}{↓~1.80}  & 78.40 & \textbf{80.40} & \textcolor{darkgreen}{↑~2.00}  & \textcolor{darkgreen}{↑~1.00} \\
\bottomrule
\end{tabular}}
\caption{Hallucination detection accuracy (\%) results on the three hallucination detection datasets.
Each cell shows the baseline performance, \ourname{} performance, and the delta. Delta values are colored: \textcolor{darkgreen}{green} for improvement, \textcolor{red}{red} for decline.}
\label{tab:nli_our_results}
\end{table*}

\cref{tab:nli_our_results} presents the results in terms of hallucination detection accuracy of the baseline (non-instructed) approach versus \ourname{} approach.
We observe a consistent performance improvement on the \textbf{ClaimVerify} and \textbf{LFQA} datasets across both standard LLMs and reasoning models—except for \texttt{Gemini-2.5-Pro} on the LFQA dataset, where performance did not improve.
For the \textbf{TofuEval} dataset, results differ between model types. Standard LLMs exhibit a performance drop relative to the baseline, whereas reasoning models show a clear improvement under \ourname{}. Overall, averaged across all models and datasets, the average accuracy gain using \textbf{\ourname{}} over the instruction-free \textbf{baseline} for the LRMs is 3.76 points. This indicates that instructing a model to make a comprehensive and systematic reasoning for an entailment decision improves the performance on NLI tasks.
Additionally, \ourname{} improvement in LRMs is twice as high as on standard LLMs. 
This suggests that reasoning models, trained to better execute reasoning steps, are more capable of following our structured and comprehensive instructions. 

The comparison between the two instruction-based reasoning approaches (\ourname{} and QA-based) and the baseline is presented in Appendix~\cref{tab:nli_full_comparison}. Both instruction-based methods lead to improved model performance, demonstrating that while self-reasoning capabilities in LRMs are valuable, explicitly guiding LRMs through a structured and principled reasoning process may further enhance their effectiveness. Additional details and insights can be found in Appendix~\ref{app:nli_comparison}.

\subsection{Ablation Study}
\label{subsec:ablations}
\begin{table*}[t]
\centering
\adjustbox{max width=\textwidth}{%
\small
\begin{tabular}{lccc}
\toprule
\textbf{Method} & \textbf{ClaimVerify} & \textbf{LFQA} & \textbf{TofuEval} \\
\midrule
Baseline                 & 71.00  & 82.62  & 68.75 \\
+ Decomposition          & 71.12  & 80.50  & 68.25 \\
+ 3-Way Classification   & 73.12  & 79.50  & \textbf{72.25} \\
+ Attribution     & \textbf{74.50} & \textbf{83.12} & 71.62 \\
\bottomrule
\end{tabular}}
\caption{Average accuracy (\%) across all models on each dataset after incrementally adding components of \ourname{} framework.}
\label{tab:nli_avg_ablations}
\end{table*}

We perform an ablation study to evaluate the individual contribution of each component in the \ourname{} process. First, we assess the impact of the \textbf{decomposition} step. In this setup, models are instructed to break down the claim into sub-claims, classify each as \emph{supported} or \emph{not supported}, and then infer whether the claim contains hallucinations based on the sub-claim classifications.

Next, we evaluate the effect of using \textbf{3-way entailment classification}. In this setup, the \emph{not-supported} category is further split into \emph{neutral} and \emph{contradiction}. Therefore, in the entailment decision classification, the model is instructed to classify each sub-claim in one of those three options.
We then test the impact of \textbf{attribution} component. In this setup, the model is instructed to identify supporting or contradicting evidence in the source for each sub-claim, if such evidence exists.
We evaluate the ablations across the three datasets using the eight models from the main experiments in \S\ref{sec:nli_experiment}. Due to computational cost, we sample 100 examples per dataset.

In Table \ref{tab:nli_avg_ablations}, we present the average accuracy across all eight models. The results indicate that the decomposition instruction yields only marginal improvements, and in some cases, even leads to decreased performance. However, we observe that explicitly distinguishing between \emph{neutral} and \emph{contradiction} labels leads to an average improvement of nearly $1$ point in accuracy. We hypothesize that the demand for fine-grained examination of the source, particularly for the distinction between \emph{neutral} and \emph{contradiction}, encourages the model to focus on more nuanced details, leading to better performance.

Additionally, as the last component of the ablation, when the instruction includes the \emph{attribution} step, performance consistently surpasses the baseline, with an average gain of $2.29$ points. Therefore, we suggest that requiring models to support their predictions with explicit evidence leads to more sound decision-making and improved performance.

Overall, the ablation findings highlight the value of the different components of \ourname{} approach and the contribution of 3-way classification and attribution steps in \ourname{}. The complete ablation results are provided in Table~\ref{tab:nli_ablations} in Appendix~\ref{app:ablations}.

\section{Human Analysis of Reasoning Quality}
\label{sec:analysis}

\subsection{Setup}
The proposed evaluation metrics, as explained in \cref{sec:metrics}, are instruction-agnostic; that is, they can be used to evaluate entailment reasoning for any instruction- and non-instruction-based reasoning process. Therefore, we also evaluate model reasoning quality under both the baseline and \ourname{} approaches.\footnote{For adjusting to other instruction-based reasoning see Appendix \ref{app:analysis}.}
Since LRMs reasoning steps are expressed in natural language---and we did not constrain the output to a specific format---we conducted a manual analysis over 200 instances. Two of the authors manually identified and evaluated the reasoning steps according to our proposed metrics.

We focus on two reasoning models, \texttt{QwQ-32B-Preview} and \texttt{DeepSeek-R1}.\footnote{\texttt{O4-mini} and \texttt{Gemini-2.5-Pro} are excluded, as their APIs do not expose intermediate reasoning tokens.} For these models, we analyze reasoning behavior on two datasets: \textbf{ClaimVerify} and \textbf{TofuEval}.
In this setup, we randomly sampled 20 instances from \textbf{ClaimVerify} and \textbf{TofuEval} datasets, and manually analyzed model behavior across the Baseline and \ourname{} settings mentioned above.
The average results over both datasets are presented in \cref{tab:analysis_average}. Separate results for \textbf{ClaimVerify} and \textbf{TofuEval} are in \cref{app:analysis} in Table \ref{tab:analysis_claimverify} and Table \ref{tab:analysis_tofueval}, respectively. As a reference, we apply the few-shot learning setting of DecompScore \citep{wanner-etal-2024-closer} and manually analyze its outputs. The number of facts in DecompScore output serves as the estimated number of gold neo-Davidsonian atomic units. Additional details on this evaluation are provided in Appendix~\ref{app:analysis}. This result in a total of 200 annotated examples.\footnote{2 datasets $\times$ 20 instances $\times$ ( 2 LRMs $\times$ ( Baseline $+$ \ourname{}) $+$ DecompScore) ) $=200$ .}

\subsection{Insights}
\label{subsec:overall_insights}

\begin{table*}[t]
\centering
\small	
\adjustbox{max width=\textwidth}{%
\begin{tabular}{llccc|cc|c}
\toprule
\textbf{Method} & \textbf{Model} 
& \multicolumn{3}{c|}{\textbf{Decomposition}} 
& \multicolumn{2}{c|}{\textbf{Fact Attribution \& Entailment}} 
& \multirow{2}{*}{\textbf{Aggregation}} \\
\cmidrule(lr){3-5} \cmidrule(lr){6-7}
& & \textbf{Atomicity} & \textbf{Soundness} & \textbf{Completeness} 
  & \textbf{Attribution} & \textbf{Entailment}  \\
\midrule
Baseline & DeepSeek-R1 & 1.55 & 0.97 & 0.90 & 0.72 & 0.85 & 1.00 \\
                               & QwQ-32B-Preview & 1.67 & 0.98 & 0.92 & 0.68 & 0.90 & 1.00 \\ \midrule
\ourname{}                            
& DeepSeek-R1 & 2.97 & 0.96 & 0.92 & 0.97 & 0.90 & 1.00 \\
& QwQ-32B-Preview & 2.95 & 0.98 & 0.95 & 0.98 & 0.99 & 1.00 \\ \midrule
Decompscore & QwQ-32B-Preview & 4.47 & 0.98 & 0.95 & -- & -- & -- \\
\bottomrule
\end{tabular}}
\caption{LRMs Reasoning Analysis – Average across ClaimVerify and TofuEval Datasets (sampled subset). The columns present the metrics, categorized according to the three \ourname{} components. The top rows show the results for the baseline approach. The second section shows the results for \ourname{} (our approach). The last row presents the Decompscore prompt values for the decomposition metrics.}
\label{tab:analysis_average}
\end{table*}

In terms of \textit{atomicity}, we find that even when models are not explicitly instructed to decompose the hypothesis, they occasionally do so. Nevertheless, \ourname{} approach consistently yields higher atomicity compared to the baseline, indicating that models generate finer-grained sub-claims when guided by \ourname{}. 
When comparing the atomicity of \ourname{} with DecompScore, we find that there is much room for improvement in terms of the granularity of the decomposition. This may be attributed to two factors: (1) \ourname{} decomposition is used as an intermediate step towards another goal, which may be less precise, and (2) the few-shot format employed in DecompScore improves decomposition quality. We leave the atomicity improvement for future work.

As explained in \cref{sec:metrics}, when the \emph{atomicity} value is high, there is a risk of hallucinating or omitting information from the original claim. However, with a low \emph{atomicity} value, the sub-claims are longer, require the attribution to be more extensive, and the entailment decision becomes complex.

In contrast, the \emph{soundness} achieved using \ourname{} is quantitatively similar to that achieved using the baseline approach. Additionally, the \emph{completeness} of \ourname{} is higher than that of the baseline approach, despite the increase in the \emph{atomicity} values of \ourname{}. Regarding the \textit{attribution} metric---which does not distinguish between incorrect and missing attributions---we observe that even in the baseline condition, models frequently provide attribution during their reasoning. However, when explicitly instructed to do so, the attribution improves substantially. This enhancement may represent one of the key contributions of \ourname{}, as further supported by the ablation results in Section \ref{subsec:ablations}.
With respect to \textit{entailment}, \ourname{} improves the \emph{entailment} score by 5 to 9 points. This might be the direct result of a better attribution step.
Finally, for \textit{aggregation}, models perform well, with perfect alignment between sub-claim classification and final claim prediction.

In the ablation setup (\S\ref{subsec:ablations}), we observe that decomposition alone yields only limited performance improvement. Additionally, as mentioned earlier, higher \emph{atomicity} facilitates easier attribution. \ourname{}, which achieves stronger performance, also scores highly on both \emph{atomicity} and \emph{attribution}. This suggests that the combination of decomposition and attribution steps during reasoning are key contributors to improving NLI performance through comprehensive and systematic reasoning. 

\section{Related Work}
\label{sec:related_work}

\paragraph{Chain-of-Thought (CoT) and Long-CoT.} 
Our work treats hallucination detection in generated text as a reasoning task, guiding CoT reasoning \citep{wei2022chain} to perform hallucination detection in an NLI fashion via decomposition, attribution, and aggregation.
Specifically, we focus on long-CoT reasoning produced by Large Reasoning Models (LRMs), where the model is prompted to accomplish multiple subtasks across a single long reasoning chain. 
This approach has proven useful in a variety of other domains that require decomposed and symbolic reasoning, such as math and coding \citep{o1openai, deepseek_v3}, with long CoTs generally following a search procedure for verification, decomposition, and backtracking \citep{marjanovic2025deepseek, gandhi2025cognitive}.
Unlike past work that has focused on applying LRMs and developing metrics for evaluating reasoning steps (e.g.  groundedness and efficiency), largely for domains like math or diagnostics \citep{lee2025evaluatingstepbystepreasoningtraces,qiu2025quantifyingreasoningabilitiesllms,chen2025spcevolvingselfplaycritic}
our work is among the first to explore long reasoning in hallucination detection, where we introduce both metrics and methods to guide and improve reasoning.

\paragraph{Hallucination Detection.}
Hallucinations---i.e. outputs that are either not faithful to the given source or contain information not grounded in any known input---occur across a wide range of generative tasks, including summarization, question answering, general text generation, and vision tasks \citep{ji2023survey}. 
Past work has addressed hallucination detection in a variety of settings \citep{Shuster2021RetrievalAR,Manakul2023SelfCheckGPTZB,Bang2023AMM,min-etal-2023-factscore} and has included training models to detect hallucinations \citep{Orgad2024LLMsKM, niu2024ragtruth, mishra2024finegrained} or to correct detected hallucinations \citep{mishrafine}, and intervening on model representations to reduce hallucination \citep{Liu2024ReducingHI}.

\paragraph{NLI Approaches.}
More closely related to our work are efforts like WiCE~\citep{kamoi2023wice} and FActScore~\citep{min-etal-2023-factscore}, and Molecular Facts \citep{gunjal-durrett-2024-molecular}, which decompose claims into sub-claims with a view to verifying claim factuality.
Our work differs from such approaches along several axes; first, unlike these approaches---which introduce decomposition methods as opposed to approaches to attribution---we go a step further by instructing the model to also find supporting or contradicting evidence for each atomic sub-claim. Additionally, in contrast to that prior work, we adopt the three-way entailment classification (entailed, contradicted, and neutral) and not the `partial-correct' class, which does not reveal the real entailment status (either neutral or contradictory).
Similarly, we treat aggregation differently from past work like WiCE, following a more logic-based NLI definition, while past work averages across claims.
Moreover, past work has focused on developing independent pieces of a verification pipeline, i.e. decomposition, attribution/entailment, or aggregation modules.
In contrast, we propose a solution in which all these steps are performed within the model's thinking step without the need of a special training for this task.

\section{Conclusion}
In this work, we leverage the explicit reasoning capabilities of LLMs, particularly Large Reasoning Models (LRMs), by providing them specific principled guidance on how to reason for entailment classification. Proposing the \ourname{} reasoning scheme, along with corresponding assessment metrics, we show that such guidance indeed improves both bottom-line entailment performance as well as reasoning quality. Future work may further investigate principled entailment reasoning by large models for additional settings and data types, as well as their potential utility for downstream tasks, like revisions and editing, and for explaining and justifying entailment decisions to humans.

\section*{Limitations}
While our work presents a structured approach for reasoning-based hallucination detection and introduces novel evaluation metrics, it has several limitations.

First, our manual reasoning analysis was conducted on a subset of datasets due to time constraints. Although it provides valuable insight into how models reason with and without instruction, a broader dataset-level evaluation would help to generalize these findings.

Second, \ourname{} uses significantly more tokens during inference. While this yields more interpretable and accurate decisions, it also increases computational cost. Future work may explore ways to balance reasoning depth with efficiency.

\section*{Ethical Considerations}
Hallucination detection plays a key role in fostering user trust in large language models (LLMs). While \ourname{} improves hallucination detection performance, it is important to acknowledge that it is not infallible. In particular, there are cases where the model incorrectly classifies a hallucinated claim as \emph{supported} by the source. This may lead users to place trust in outputs that contain factual errors. As such, systems that integrate \ourname{} method should be transparent about its limitations and avoid presenting outputs as unquestionably reliable. Therefore, we encourage responsible deployment that includes user-facing disclaimers.


\appendix
\newpage

\vspace{0.5em}

The following appendix is structured as follows:

\begin{itemize}

    \item \textbf{Appendix~\ref{app:nli_experiment}} contains supplementary details and results on the NLI experiments.
  \item \textbf{Appendix~\ref{app:ablations}} contains additional ablation results.
  \item \textbf{Appendix~\ref{metrics_sec_app}} contains additional details regarding the use of the evaluation metrics for QA-based instructions.
  \item \textbf{Appendix~\ref{app:analysis}} contains additional experimental analysis, including both decomposition and a manual analysis.
  \item \textbf{Appendix~\ref{app:prompts}} contains the prompts used within our experiments. 
\end{itemize}

\section{NLI Experiments}
\label{app:nli_experiment}

This section presents additional supplementary details and results related to the NLI experiments. Subsection~\ref{app:nli_datasets} offers further information about the datasets used, while Subsection~\ref{app:nli_comparison} compares our approach with the QA-based method.

\subsection{Datasets}
\label{app:nli_datasets}
We evaluate \ourname{} process for hallucination detection using datasets from the Natural Language Inference (NLI) task, where each instance includes: (1) a \textit{premise} — a reliable source document, (2) a \textit{hypothesis} — a text segment generated by a large language model, and (3) a \textit{label} - indicating whether the hypothesis is supported by the premise.

\paragraph{ClaimVerify.} For the fact verification domain, we use the ClaimVerify dataset \cite{liu-etal-2023-evaluating}. ClaimVerify assesses the factual accuracy of responses from four generative search engines in answering user queries. Each instance includes a sentence from a generated response and its associated source document, annotated to indicate whether the sentence is fully supported by the cited source. We selected this dataset due to its diversity: it contains generations from four different models, might capturing a wide range of behaviors and hallucinations.

\paragraph{LFQA-Verification.} In the question answering domain, we evaluate on the LFQA-Verification dataset \cite{chen2023understanding}. LFQA-Verification consists of responses generated by LLMs to questions from the ELI5 dataset \cite{fan-etal-2019-eli5}. The models generate responses based on documents retrieved either by humans, retrieval models, or selected at random. Human annotators label each sentence in the generated responses as \textit{supported}, \textit{partially supported}, or \textit{not supported}. For consistency across datasets, our experiment combines the partially supported and not supported labels into a single \textit{not supported} label.

\paragraph{TofuEval.} For summarization, we use the TofuEval dataset \cite{tang-etal-2024-tofueval} based on the MediaSum benchmark \cite{zhu-etal-2021-mediasum}. TofuEval targets factual consistency in dialogue summarization, focusing on interview transcripts from MediaSum. It includes topic-focused summaries generated by six different LLMs, with sentence-level factual consistency annotations provided by linguists. The dataset's coverage across multiple models contributes valuable diversity to the evaluation.

The datasets described above contain thousands of samples. Due to the high computational cost of running inference on LRMs, we sample 500 instances from each dataset (sample IDs will be released upon acceptance). Since many prior works report only the balanced accuracy \cite{5597285}, a metric that adjusts class imbalance, for the hallucination detection task \cite{10.1162/tacl_a_00453, tang-etal-2024-minicheck, tang-etal-2024-tofueval, paudel2025hallucinothallucinationdetectioncontext}, we adopt a balanced sampling strategy. Specifically, we randomly sample 250 supported and 250 not-supported instances from each dataset.
All the datasets have been imported via LLM-AggreFact collection, available on HuggingFace \citep{tang-etal-2024-minicheck}

\paragraph{Binary Classification.} Most recent hallucination-detection datasets adopt a binary classification setup, labeling each claim as either \textit{supported} or \textit{not supported}. This mirrors real-world applications, where users are typically concerned with whether to trust a model's output. Therefore, in this work, we also focus on binary hallucination classification: determining whether a generated text (i.e., a claim) contains hallucinations, without distinguishing whether the hallucination is either a `contradiction' or `neutral' relative to the source. However, since \ourname{} framework does support fine-grained distinctions between contradiction and neutrality, it may offer additional benefits for other downstream applications. We leave this exploration for future work.

\subsection{NLI Methods Comparison}
\label{app:nli_comparison}

\begin{table*}[t]
\centering
\adjustbox{max width=\textwidth}{%
\begin{tabular}{l|ccc|ccc|ccc}
\toprule
\textbf{Model} 
& \multicolumn{3}{c|}{\textbf{ClaimVerify}} 
& \multicolumn{3}{c|}{\textbf{LFQA}} 
& \multicolumn{3}{c}{\textbf{TofuEval}} \\
& Baseline & QAs & \ourname{} 
& Baseline & QAs & \ourname{} 
& Baseline & QAs & \ourname{} \\
\midrule
MiniCheck & 60.20 & -- & -- & 55.60 & -- & -- & 66.20 & -- & -- \\
\midrule
Qwen-Plus & 71.00 & 73.20 & \textbf{74.40} & 79.60 & 78.80 & \textbf{81.00} & \textbf{78.60} & 76.20 & 71.40 \\
DeepSeek-V3 & 66.60 & 69.80 & \textbf{73.40} & 80.60 & 82.60 & \textbf{84.00} & \textbf{77.80} & 77.60 & 77.20 \\
GPT-4o-mini & 71.40 & 65.00 & \textbf{73.80} & 77.60 & 75.00 & \textbf{83.20} & \textbf{79.00} & 65.80 & 78.00 \\
Gemini-2.0-Flash & 68.00 & 69.80 & \textbf{75.00} & 78.20 & \textbf{80.60} & \textbf{80.60} & \textbf{78.60} & 78.40 & 78.20 \\
\midrule
QwQ-32B-Preview & 67.40 & 71.80 & \textbf{72.40} & 79.80 & 81.40 & \textbf{82.40} & 70.22 & 78.60 & \textbf{79.80} \\
DeepSeek-R1 & 69.60 & 70.40 & \textbf{75.60} & 80.60 & 80.40 & \textbf{84.40} & 71.23 & 72.60 & \textbf{77.00} \\
O4-mini & 73.20 & 74.00 & \textbf{80.20} & 85.80 & 86.20 & \textbf{86.80} & 80.20 & 81.20 & \textbf{81.60} \\
Gemini-2.5-Pro & 73.40 & 75.60 & \textbf{76.20} & 85.80 & \textbf{87.00} & 84.00 & 78.40 & 80.20 & \textbf{80.40} \\
\midrule
\textbf{Average (LRMs)} & 70.90 & 72.95 & \textbf{76.10} & 83.00 & 83.75 & \textbf{84.40} & 75.01 & 78.15 & \textbf{79.70} \\
\bottomrule
\end{tabular}}
\caption{Comparison of performance across three datasets for various models using different reasoning strategies. Each cell shows accuracy (\%); the best value per row is bolded.}
\label{tab:nli_full_comparison}
\end{table*}

We conducted a comparison of two instruction-based reasoning approaches: QA-based approach, and \ourname{} approach. \ourname{} is descrin details in Section \ref{sec:definition}. In the QA-based approach, we instruct the model to first generate questions on the claim. Then, the model is guided to answer the questions based on the claim and based on the source, separately. Finally, the model is instructed to compare the answers and consequently decide on the final decision of the claim. That is, if a claim's answer is not equivalent to a source's answer, the information from the source that is represented by this question-and-answer is not faithful to the source.  The full prompts are presented in Appendix \ref{app:prompts}.

The results for each approach, along with the baseline results, are presented in Table~\ref{tab:nli_full_comparison}. 
The comparison was conducted across all eight models, with the full results shown in Table~\ref{tab:nli_full_comparison}. However, given that the primary focus of this paper is on LRMs, the following analysis will emphasize results from LRMs specifically.
We find that \ourname{} approach achieves the highest average performance on the \textbf{ClaimVerify} and \textbf{TofuEval} datasets, and \textbf{LFQA} dataset, with an overall average accuracy of $80.7\%$. The QA-based method ranks second across all three datasets, with an overall average accuracy of $78.28\%$. The baseline approach performs the worst in all datasets, with an average accuracy of $76.3\%$.
These findings indicate that while self-reasoning capabilities in LRMs are beneficial, explicitly guiding LRMs to reason in a structured and principled manner may further enhance their performance.

\section{Additional Ablation Results}
\label{app:ablations}
\begin{table*}[t]
\centering
\begin{tabular}{llrrr}
\toprule
\textbf{Model} & \textbf{Method} & \textbf{ClaimVerify} & \textbf{LFQA} & \textbf{TofuEval} \\
\midrule
\multirow{4}{*}{Qwen-Plus}
& Baseline        & 68.00 & 83.00 & 66.00 \\
& + Decomposition & 67.00 & 81.00 & 61.00 \\
& + 3 way         & \textbf{77.00} & 76.00 & \textbf{74.00} \\
& + Attribution   & 74.00 & \textbf{86.00} & 65.00 \\
\midrule
\multirow{4}{*}{DeepSeek-V3}
& Baseline        & 70.00 & 83.00 & 69.00 \\
& + Decomposition & 72.00 & 83.00 & \textbf{70.00} \\
& + 3 way         & 74.00 & 83.00 & 69.00 \\
& + Attribution   & \textbf{77.00} & \textbf{86.00} & \textbf{70.00} \\
\midrule
\multirow{4}{*}{GPT-4o-mini}
& Baseline        & 70.00 & \textbf{84.00} & \textbf{71.00} \\
& + Decomposition & 68.00 & 75.00 & 65.00 \\
& + 3 way         & 66.00 & 72.00 & 66.00 \\
& + Attribution   & \textbf{73.00} & 81.00 & 66.00 \\
\midrule
\multirow{4}{*}{Gemini-2.0-Flash}
& Baseline        & 71.00 & \textbf{84.00} & 66.00 \\
& + Decomposition & 70.00 & 76.00 & 68.00 \\
& + 3 way         & 70.00 & 78.00 & \textbf{78.00} \\
& + Attribution   & \textbf{75.00} & 81.00 & \textbf{78.00} \\
\midrule
\multirow{4}{*}{QwQ-32B-Preview}
& Baseline        & 70.00 & 80.00 & 68.00 \\
& + Decomposition & 73.00 & \textbf{85.00} & 72.00 \\
& + 3 way         & \textbf{74.00} & 79.00 & \textbf{76.00} \\
& + Attribution   & 73.00 & 83.00 & 70.00 \\
\midrule
\multirow{4}{*}{DeepSeek-R1}
& Baseline        & 71.00 & \textbf{80.00} & 69.00 \\
& + Decomposition & 74.00 & \textbf{80.00} & \textbf{73.00} \\
& + 3 way         & \textbf{76.00} & \textbf{80.00} & 72.00 \\
& + Attribution   & 73.00 & 77.00 & \textbf{73.00} \\
\midrule
\multirow{4}{*}{O4-mini}
& Baseline        & 74.00 & 84.00 & \textbf{71.00} \\
& + Decomposition & 72.00 & 86.00 & 70.00 \\
& + 3 way         & 74.00 & \textbf{87.00} & \textbf{71.00} \\
& + Attribution   & \textbf{75.00} & \textbf{87.00} & \textbf{71.00} \\
\midrule
\multirow{4}{*}{Gemini-2.5-Pro}
& Baseline        & 74.00 & 83.00 & 70.00 \\
& + Decomposition & 73.00 & 78.00 & 67.00 \\
& + 3 way         & 74.00 & 81.00 & 72.00 \\
& + Attribution   & \textbf{76.00} & \textbf{84.00} & \textbf{80.00} \\
\bottomrule
\end{tabular}
\caption{Full ablation results across all models. We randomly sampled 100 instances from each dataset.}
\label{tab:nli_ablations}
\end{table*}

This section presents additional ablation results that were not presented from the main paper due to space limitations. The full ablation results for \ourname{} process across all eight models are presented in Table~\ref{tab:nli_ablations}.

One notable observation is that the decomposition step on its own often leads to a decrease in performance. This is likely because LLMs are not explicitly trained to perform atomic-level decomposition, and prompting them to do so may lead to confusion or misinterpretation of the task. In contrast, we find that distinguishing between the \emph{Contradiction} and \emph{Neutral} classes improves performance in half of the models evaluated. Similarly, the attribution step also improves the performance in half of the cases.
These findings suggest that the comprehensiveness of \ourname{}---particularly the inclusion of fine-grained 3-way entailment classification and attribution---contributes positively to the quality of reasoning in the entailment task.

\section{Using Metrics for QA-based Instructions}
\label{metrics_sec_app}
In Section~\ref{sec:metrics}, we argue that our proposed evaluation metrics are instruction-agnostic, i.e., they can evaluate reasoning for NLI regardless of the reasoning process followed. For both \ourname{} flow and instruction-free reasoning, we explain in the paper how to apply these metrics. However, applying those metrics to QA-based instructions requires some clarification.

In the QA-based setting, the model is instructed to generate questions based on the claim, answer them using the claim itself, and then answer them again using the source document. The model then compares these two sets of answers to assess the correctness of each sub-claim and, by extension, the entire claim.

The proposed metrics can be naturally adapted to this process as follows: the generated questions correspond to the decomposition step; the model’s answers from the source act as the attribution; the comparison between claim-based and source-based answers serves as the entailment classification; and the final judgment, whether all answers align, constitutes the aggregation step.

\section{Additional Experimental Analysis}
\label{app:analysis}

This section presents additional experimental analysis, including the decomposition-based experiment (Subsection~\ref{decomposiotion_subsec_ap}) and further manual analysis (Subsection~\ref{app:analysis_results}).

\subsection{Decomposition}
\label{decomposiotion_subsec_ap}
In addition to the analysis on baseline and \ourname{} approaches, we wanted to compare the atomicity values with the number of 'gold' atomicity. However, since it's time-consuming, we did the same as \cite{wanner-etal-2024-closer} and prompted a model, with a few-shot examples for neo-Davidsonian samples to provide a new-Davidsonian decomposition. We believe that since this is the only task of this prompt, compared to \ourname{}, the output should be much closer to the gold neo-Davidsonian decomposition.
For this, we used the \texttt{QwQ-32B-Preview} model and instructed him to do the decomposition. Then, we manually evaluate its output on the \textit{atomicity}, \textit{soundness}, and \textit{completeness}. However, the main comparison here is for the atomicity compared to the atomicity of the NLI instructions.

\subsection{Manual Analysis}
\label{app:analysis_results}
\begin{table*}[t]
\centering
\adjustbox{max width=\textwidth}{%
\begin{tabular}{llcccccc}
\toprule
\textbf{Method} & \textbf{Model} & \textbf{Atomicity} & \textbf{Soundness} & \textbf{Completeness} & \textbf{Entailment} & \textbf{Attribution} & \textbf{Aggregation} \\
\midrule
Baseline & DeepSeek-R1 & 1.55 & 0.97 & 0.95 & 0.95 & 0.72 & 1.00 \\
                               & QwQ-32B-Preview & 1.75 & 1.00 & 0.90 & 0.92 & 0.82 & 1.00 \\ \hline
\ourname{}                           
& DeepSeek-R1 & 2.65 & 0.97 & 0.95 & 0.87 & 0.95 & 1.00 \\
& QwQ-32B-Preview & 2.85 & 0.98 & 1.00 & 0.99 & 1.0 & 1.00 \\ \hline
Decompscore & QwQ-32B-Preview & 4.30 & 0.98 & 1.00 & -- & -- & -- \\
\bottomrule
\end{tabular}}
\caption{Reasoning Analysis – ClaimVerify Dataset (sampled subset)}
\label{tab:analysis_claimverify}
\end{table*}

\begin{table*}[t]
\centering
\adjustbox{max width=\textwidth}{%
\begin{tabular}{llcccccc}
\toprule
\textbf{Method} & \textbf{Model} & \textbf{Atomicity} & \textbf{Soundness} & \textbf{Completeness} & \textbf{Entailment Accuracy} & \textbf{Attribution} & \textbf{Aggregation} \\
\midrule
Baseline & DeepSeek-R1 & 1.55 & 0.97 & 0.85 & 0.75 & 0.73 & 1.00 \\
         & QwQ-32B-Preview & 1.60 & 0.97 & 0.95 & 0.88 & 0.55 & 1.00 \\ \hline
\ourname{}     
         & DeepSeek-R1          & 3.30 & 0.96 & 0.90 & 0.93 & 1.00 & 1.00 \\
         & QwQ-32B-Preview & 3.05 & 0.98 & 0.90 & 0.99 & 0.97 & 1.00 \\ \hline
Decompscore & QwQ-32B-Preview & 4.65 & 0.98 & 0.90 & -- & -- & -- \\
\bottomrule
\end{tabular}}
\caption{Reasoning Analysis – TofuEval Dataset (sampled subset)}
\label{tab:analysis_tofueval}
\end{table*}

The manual analysis results for ClaimVerify are Table \ref{tab:analysis_claimverify}.
The manual analysis results for TofuEval are Table \ref{tab:analysis_tofueval}.

\section{Description of Prompts}
\label{app:prompts}

This section contains the prompts used within our experiments. Particularly, \begin{inparaenum}[(i)] \item 
    Subsection~\ref{hallucination_detec_prompt} contains the hallucination detection prompts, \item Subsection~\ref{Decomposition_prompt} contains the decomposition prompts, \item Subsection~\ref{CoReference_prompts} contains the co-reference prompts, and \item Subsection~\ref{ablation_prompts} contains the ablation prompts.
\end{inparaenum} 

\tcbset{
  promptstyle/.style={
    fontupper=\small	
  }
}

\subsection{Hallucination Detection Prompts}
\label{hallucination_detec_prompt}

We present here the prompts used for the hallucination detection task. To ensure consistency with prior work, we adopt the baseline prompt from \citet{tang-etal-2024-minicheck}, as presented in Prompt 1.1. For the \textit{<specific instructions for each method>}, there is a variant for each instruction approach. For the baseline approach, it is left empty.

For standard LLMs, we augment the prompt with chain-of-thought (CoT) reasoning \cite{wei2022chain} by inserting the phrase ``think step by step'' as the \textit{<instruction for chain of thought>}.
The decomposition-based prompt and QA-based prompt variants for the \textit{<specific instructions for each method>} are included in Prompts 1.2 and 1.3, respectively.
The instructions version for \ourname{} is shown in Prompt 1.4, while an example of Davidsonian-inspired decomposition appears in Prompt 1.5. Prompt.

\subsection{Decomposition}
\label{Decomposition_prompt}
We note that although we instruct the model to decompose the hypothesis into atomic facts, our goal was not to optimize decomposition quality, and in practice, the models do not always succeed in producing atomic facts. Therefore, we refer to this step as a decomposition into smaller sub-claims, rather than strictly atomic ones.

\subsection{Co-Reference Between Atomic Facts}
\label{CoReference_prompts}

\citet{gunjal-durrett-2024-molecular} highlight that decomposing a text segment into atomic facts may not be sufficient for detecting hallucinations. One key reason is that contradictions can arise not from individual facts themselves, but from their \textit{co-reference}. That is, two atomic facts may each be individually entailed by the premise, yet their combination, through shared referents, can result in a contradiction.

For example, consider the premise: \textit{``Ann Jansson is a Swedish former footballer. Another Ann Jansson, a racewalking athlete, won a medal at the European Athletics Championships.''} Now consider the hypothesis: \textit{``Ann Jansson is a Swedish former footballer who won the European Athletics Championships.''}. When decomposed, the hypothesis yields two sub-facts: (1) \textit{``Ann Jansson is a Swedish former footballer''} and (2) \textit{``Ann Jansson won a medal at the European Athletics Championships.''}. Both sub-facts are individually entailed by the premise. However, the co-reference between the two distinct individuals named ``Ann Jansson'' introduces a contradiction relative to the premise.

To address this, we instructed the model to also evaluate whether co-reference across sub-facts introduces a contradiction.
In the manual analysis, we found that while models were capable of executing this step, they never identified an actual contradiction arising from co-reference. Therefore, we did not explicitly incorporate this property into the main evaluation framework presented in the paper.

\subsection{Ablation Prompts}
\label{ablation_prompts}
For the ablations, which are described in Section \ref{subsec:ablations}, the \textit{baseline} approach uses Prompt 1.1. The prompt for the \textit{decomposition} approach, which is inspired by Davidsonian semantics, is Prompt 2.1. For the \textit{3-way} approach, we instruct the model according to Prompt 2.2. The instruction for the \textit{attribution} approach is the same as Prompt 1.4.

\newpage
\onecolumn

\vspace{1em}
\begin{tcolorbox}[promptstyle, title=\texttt{Prompt 1.1: NLI Baseline}]
Determine whether the provided claim is consistent with the corresponding document. Consistency in this context implies that all information presented in the claim is substantiated by the document. If not, it should be considered inconsistent.
\newline \newline
Document: \texttt{\{\{document\}\}} \\
Claim: \texttt{\{\{claim\}\}}
\newline \newline
\textit{\textless specific instructions for each method\textgreater} \newline \newline
Conclude your response with either ``yes'' (the claim is supported) or ``no'' (the claim is not supported).
\newline\newline
\textit{\textless instruction for chain of thought\textgreater}
\end{tcolorbox}

\begin{tcolorbox}[promptstyle, title=\texttt{Prompt 1.2: QA-Based Instructions}]
Follow the steps below to guide your assessment:
\begin{enumerate}[leftmargin=1.5em]
    \item Generate questions based on the claim.
    \item Answer those questions based on the document and on the claim separately.
    \item Check if the documents' answers and the claims' answers are similar.
    \item Make a final decision based on your analysis.
\end{enumerate}
\end{tcolorbox}

\begin{tcolorbox}[promptstyle, title=\texttt{Prompt 1.3: Decomposition-Based Instructions}]
Follow the steps below to guide your assessment:
\begin{enumerate}[leftmargin=1.5em]
    \item Split the claim into separate sentences.
    \item Split each sentence into a few parts. Each part should contains a different topic of the sentence. 
    For example, for the claim: ``A blue motorcycle parked by paint-chipped doors.'', its parts are:
    - ``A blue motorcycle parked by doors'' \newline
    -``A motorcycle parked by paint-chipped doors''
    \item For each part, evaluate its support within the document.
    \item Make a final decision based on your analysis.
\end{enumerate}

\end{tcolorbox}

\begin{tcolorbox}[promptstyle, title=\texttt{Prompt 1.4: Comprehensive Reasoning Instructions}]
Follow the steps below to guide your assessment:
\begin{enumerate}[leftmargin=1.5em]
    \item Split the claim into separate sentences.
    
    \item Decompose each sentence into its atomic components. \newline An atomic proposition is a statement that: \newline
    (i) has a truth value verifiable against the document, and \newline
    (ii) cannot be broken down further into smaller factual units with distinct truth values. \newline
    \texttt{ \{\{example\}\} }
    
    \item For each atomic component, evaluate its support within the document. \newline
    - If supported, identify the exact phrase in the document that confirms it. \newline
    - If contradicted, cite the phrase that disproves it. \newline
    - If neither supported nor contradicted, mark it as a neutral component.

    \item Evaluate combinations of atomic facts. \newline
    - If a combination is supported or contradicted, provide the source phrase(s) for this judgment.

    \item Make a final decision based on your analysis: \newline
    - If there is at least one contradiction or neutral component, the claim is not supported. \newline
    - If all components are entailed by the document, the claim is supported.
\end{enumerate}
\end{tcolorbox}

\begin{tcolorbox}[promptstyle, title=\texttt{Prompt 1.5: Davidsonian-Inspired Decomposition Example}]
For example, for the claim: for the claim: `A blue motorcycle parked by paint chipped doors.', its atomic facts are: `the motorcycle is blue', `the motorcycle is parked', `the doors are paint', `the door is paint chipped', `the motorcycle is next to the doors'.
\end{tcolorbox}

\begin{tcolorbox}[promptstyle, title=\texttt{Prompt 2.1: Davidsonian-inspired Decomposition Instructions}]
Follow the steps below to guide your assessment:

\begin{enumerate}[leftmargin=1.5em]
    \item Split the claim into separate sentences.
    
    \item Decompose each sentence into its atomic components. \newline
    An atomic proposition is a statement that: \newline
    (i) has a truth value verifiable against the document, and \newline
    (ii) cannot be broken down further into smaller factual units with distinct truth values.\newline
    \texttt{\{\{example\}\}}
    
    \item For each atomic component, determine whether it is supported by the document (i.e., can be inferred from the document), or not supported by the document.

    \item Make a final decision based on your analysis: \newline
    - If there is at least one contradiction or neutral component, the claim is not supported.\newline
    - If all components are entailed by the document, the claim is supported.
\end{enumerate}
\end{tcolorbox}

\begin{tcolorbox}[promptstyle, title=\texttt{Prompt 2.2: Davidsonian-inspired Decomposition Instructions}]
Follow the steps below to guide your assessment:

\begin{enumerate}[leftmargin=1.5em]
    \item Split the claim into separate sentences.
    
    \item Decompose each sentence into its atomic components. \newline
    An atomic proposition is a statement that: \newline
    (i) has a truth value verifiable against the document, and \newline
    (ii) cannot be broken down further into smaller factual units with distinct truth values.\newline
    \texttt{\{\{example\}\}}
    
    \item For each atomic component, determine whether it is supported by the document (i.e., can be inferred from the document), contradicted by the document, or neutral relative to the document.

    \item Make a final decision based on your analysis: \newline
    - If there is at least one contradiction or neutral component, the claim is not supported.\newline
    - If all components are entailed by the document, the claim is supported.
\end{enumerate}
\end{tcolorbox}

\end{document}